\documentclass[runningheads]{llncs}
\usepackage{graphicx}
\usepackage{url}
\usepackage{listings}
\begin{document}
\title{The Music Note Ontology\thanks{Copyright © 2021 for this paper by its authors. Use permitted under Creative Commons License Attribution 4.0 International (CC BY 4.0).}\textsuperscript{,}\thanks{This project has received funding from the European Union’s Horizon 2020 research and innovation programme under grant agreement No 101004746.}}
%
%
\author{Andrea Poltronieri\inst{1}\thanks{Corresponding author: andrea.poltronieri2@unibo.it}\orcidID{0000-0003-3848-7574} \and
Aldo Gangemi\inst{2}\orcidID{0000-0001-5568-2684}}
\authorrunning{A. Poltronieri and A. Gangemi}
%
\institute{LILEC, University of Bologna, Bologna, Italy \and FICLIT, University of Bologna, Bologna, Italy}

\maketitle              
\begin{abstract}
In this paper we propose the \emph{Music Note Ontology}, an ontology for modelling music notes and their realisation. The ontology addresses the relation between a note represented in a symbolic representation system, and its realisation, i.e. a musical performance. 
This work therefore aims to solve the modelling and representation issues that arise when analysing the relationships between abstract symbolic features and the corresponding physical features of an audio signal.
The ontology is composed of three different Ontology Design Patterns (ODP), which model the structure of the score (\emph{Score Part Pattern}), the note in the symbolic notation (\emph{Music Note Pattern}) and its realisation (\emph{Musical Object Pattern}).

\end{abstract}
%
%
\section{Introduction}

A \emph{music note} is defined as ``the marks or signs by which music is put on paper. Hence the word is used for the sounds represented by the notes'' \cite{grove1908}.  However, musical notation provides some general information on how to play a certain note. These indications are then enriched in the context of a performance by a large amount of information, for example, from the musician's sensitivity or the conductor's instructions. Historically, music notation has been a crucial innovation that allowed the study of music and hence the rise of musicology. Music scores were initially introduced for the primary purpose of recording music by giving other musicians the possibility of playing the same piece in turn, reproducing it. However, musical notation by definition entails expressive constraints on the description of musical content. 

Representing music involves a number of issues that are closely related to the complexity of the musical content. This can be found in every system of music representation, from music scores to their various forms of digital encoding, usually referred to as \emph{symbolic representation systems}. As Cook observed, music scores symbolise rather than represent music, as people don’t play musical rhythms as written and often they don’t play the pitches as written: that is because the notation is only an approximation \cite{cook2006}. To underline this concept, it is useful to draw a distinction between a \emph{score} and a \emph{musical object} \cite{wiggins1993}. While the former can be thought of as a set of instructions that the musician or a computer system uses to realise a piece of music, a musical object consists of the realisation of this process. In other words, musical notation precedes realisation or interpretation and defines only partially the musical object \cite{nattiez1977}. 

Music notation is not just a mechanical transformation of performance information. Performance nuances are lost when moving from performance to notation, and symbolic structure is partly lost in the translation from notation to performance \cite{dannenberg1993}. 
Moreover, in music notation some features of  musical content are completely overlooked. For example, timbre information is generally ignored and is at best provided through generic information about the instrument that is supposed to play the part. 
Many of these problems have not been overcome by the numerous systems of symbolic representation proposed in recent decades. Music representation systems (MSR) have been evaluated by Wiggins et al. \cite{wiggins1993} using a system based on two orthogonal dimensions. These two dimensions have been named as \emph{expressive completeness} and \emph{structural generality}. Expressive completeness is described as the extent to which the original musical content may be retrieved and/or recreated from its representation \cite{wiggins2009}. Structural generality, instead, refers to the range of high-level structures that can be represented and manipulated \cite{wiggins1993}. For example, raw audio is very expressive since it contains a great deal of information about a certain music performance; but a MIDI representation contains much less information, e.g. timbral features are not represented in this format. On the contrary, when evaluating structural generality, raw audio performs poorly, due to the difficulties in the extraction of structured information (e.g., tempo, chords, notes, etc.) from such a format.

Another problem in representing music is linked to the twofold nature of musical content, which contains information that can be reduced to mathematical functions, but also information related to the emotional spectrum and a wide range of psychoacoustic nuances. In fact, music is distinguished by the presence of many relationships that can be treated mathematically, including rhythm and harmony. However, elements such as tension, expectancy, and emotion are less prone to mathematical treatment \cite{dannenberg1993}. 

The limits of symbolic representations are also accentuated when the performance is considered in relation to the concept of interpretation. For each composition, several performances of the same one correspond to as many interpretations of the musical piece. These interpretations can also vary greatly from one another from agogic (note accentuation, \cite{brown1991}), timbral and dynamic viewpoints. 

Therefore, this work aims to represent musical notes both symbolically and as a realisation of the note itself. The alignment between these two representation systems can be relevant on several grounds. For example, it can allow musical analysis using a structured representation (i.e. a symbolic representation) while simultaneously taking into account all the information that is only contained in the signal (e.g. timbre information). This would allow a different level of analysis, taking into account information that is usually overlooked in music scores and symbolic representations. On the other side,  this type of representation would allow structured analysis of the music signal by having the signal information encoded in strict relation with the music score and the multiple hierarchies that the music score entails.

Furthermore, this type of representation allows the analysis of different realisations of the same note, also providing information on how a score is performed differently in different performances.

\section{Physical Features and Perceptual Features}

The features that need to be taken into account when representing music can be reduced to four main categories: tempo, pitch, timbre and dynamics. 
However, all these concepts refer to music perception. Symbolic representations, however, only represent abstractions of these features. In contrast, musical performance, as analysed in recorded form (i.e. signal representation), contains information that refers to physical characteristics. Through the analysis of these characteristics, it is in turn possible to abstract the perceptual characteristics aroused by a particular sound. However, this work aims not to analyse the perceptual aspects of music, but rather to formalise a representation that aims to express a musical note from the points of view of both scores and musical objects. 

As far as time is concerned, the main distinction that arises is that between the quantised time of a symbolic representation, and real time, which is expressed in seconds. A symbolic representation describes temporal information using predetermined durations for each note (e.g., quarter notes, eighth notes, sixteenth notes, etc.). These durations are executed in relation to a time signature, indicated by a tempo marking (e.g., slow, moderate, allegro, vivace, etc.), or by a metronome mark, which indicates the tempo measured in beats per minute (bpm). 
Generally, symbolic notations (e.g. MusicXML) use the same expedients as classical musical notation. However, some representation systems rely on real-time, such as the MIDI standard \cite{midi1996}. 

When considering frequency, the representation issue is more challenging. A sound, e.g. a note in the form of a musical note, is usually associated with a pitch. However, although pitch is associated with frequency \cite{moore2003}, it is a psychological percept that does not map straightforwardly onto physical properties of sound \cite{mclachlan2016}. The American National Standards Institute (ANSI) defines pitch as the “auditory attribute of sound according to which sounds can be ordered on a scale from low to high” \cite{ansi1973}. In fact, in many languages, the pitch is described by using terms having a spatial connotation such as “high” and “low” \cite{rusconi2006}. However, the spatial arrangement of sounds has been shown to be influenced by the listener's musical training \cite{stewart2004} and their ethnic group \cite{antovic2009}. 
The relation between pitch and frequency is evident in the case of pure tones \cite{muller2015}. For example, a sinusoid having a frequency of 440 Hz corresponds to the pitch A4. 
However, real sounds are not composed of a simple pure note with a unique and well-defined frequency. Playing a single note on an instrument may result in a complex sound that contains a mixture of different frequencies changing over time. Intuitively, such a musical tone can be described as a superposition of pure tones or sinusoids, each with its frequency of vibration, amplitude, and phase. A partial is any of the sinusoids, by which a musical tone is described \cite{muller2015}. The frequency of the lowest partial is defined as the fundamental frequency. Most instruments produce harmonic sounds, i.e. composed of frequencies close to the partial harmonics. However, some instruments such as xylophones, bells, and gongs produce inharmonic sounds, i.e. composed of frequencies that are not multiples of the fundamental. As a result, the pitch perception of these sounds is distorted for a listener who is unfamiliar with these instruments \cite{mclachlan2016}. Several approaches have been proposed to determine the perceived pitch of these sounds, based both on the periodicity of the frequencies that compose the sound \cite{schouten1938} and on spectral clues \cite{carlyon1994}. However, recent research suggests that pitch perception involves learning to recognise a sound timbre over a range of frequencies, and then associating changes in frequency with visuospatial and kinesthetic dimensions of an instrument \cite{mclachlan2016}.

On the other hand, timbre is a sound property even more difficult to grasp. ANSI defines the timbre as the ``attribute of auditory sensation, in terms of which a subject can judge that two sounds similarly presented and having the same loudness and pitch are dissimilar'' \cite{ansi1973}. However, even today, the definition of timbre is rather debated. For example, it has been defined as a “misleadingly simple and vague word encompassing a very complex set of auditory attributes, as well as a plethora of psychological and musical issues” \cite{mcadams2008}.
The main problem in defining timbre is that there is no precise correspondence with one or more physical aspects. Timbre is also challenging to abstract, as it is usually described using adjectives such as light, dark, bright, rough, violin-like, etc. Numerous studies have shown that timbre depends on a large number of factors and physical characteristics. Some research has tried to represent the perception of timbre using multidimensional scaling (e.g. \cite{grey1977}). Other studies in this area focus on the grouping of timbre into a ``Timbre Space'' \cite{mcadams2008}. 
While those studies represented timbre in terms of perceptual dimensions, others have represented timbre in terms of physical dimensions, such as the set of parameters needed for a particular synthesis algorithm \cite{dannenberg1993}. These physical characteristics concern both the time axis, such as attack time and vibrato variation and amplitude, spectral features such as spectral centroid, and harmonic features such as odd-to-even ratio and inharmonicity. An approximation of these features is provided by a model called \emph{ADSR}, which consists of the analysis of the envelope of the sound wave and the measurement of attack (A), decay (D), sustain (S) and release (R). The envelope can be defined as a smooth curve outlining the extremes in the amplitude of a waveform. The ADSR model, however, is a strong simplification and only yields a meaningful approximation for amplitude envelopes of tones that are generated by specific instruments \cite{muller2015}.

The same distinction drawn between frequency and pitch can be made for loudness and sound intensity. Loudness is a perceptual property whereby sounds are ordered on a scale from quiet to loud. Sound intensity is generally measured in dB, and expresses the sound power (expressed in Watts) in relation to the physical space in which the sound propagates (expressed in square metres) \cite{bruneau2013}. As a perceptual property, it is by its nature subjective and is closely related to loudness. However, loudness also depends on other sound characteristics such as duration or frequency \cite{muller2015}. Also, the sound duration influences perception since the human auditory system averages the effect of sound intensity over an interval up to a second. 
 
Formalizing the aforementioned perceptual concepts is beyond the scope of this paper. Instead, the purpose of this contribution is to define (some of) the attributes that can provide helpful information to describe sound material. 
The importance of this analysis is to highlight the inner complexity of the sound material, underlying the connections that can be drawn between physical aspects and perceptual components, as well as the relationships among these aspects.

\section{Related Work}

Several Semantic Web ontologies have been proposed for modelling musical data. 
The Music Theory Ontology (MTO) \cite{rashid2018} aims at modelling music theoretical concepts. This ontology focuses on the rudiments that are necessary to understand and analyse music. However, this ontology is limited to the description of note attributes (dynamics, duration, articulation, etc.) at the level of detail of a note set. This peculiarity considerably reduces the expressiveness of the model, especially in relation to polyphonic and multi-voice music.

The Music Score Ontology (Music OWL) \cite{jones2017} aims to represent similar concepts described in the Music Theory Ontology. However, it focuses on music notation, and the classes it is composed of are all aimed at representing music in a notated form, hence related to music sheets.

The Music Notation Ontology \cite{cherfi2017} is an ontology of music notation content, focused on the core ``semantic'' information present in a score and subject to the analytic process. Moreover, it models some high-level analytic concepts (e.g. dissonances), and associates them with fragments of the score.

However, all these ontologies only consider the characteristics of the musical material from the point of view either symbolic representation or music score. The aim of this paper is instead to propose a model that can be as general as possible and that can therefore represent musical notation expressed in different formats and considering at the same time scores and symbolic notations. This would allow interoperability between different notation formats and the standardisation of musical notation in a format that aims to be as expressive as possible. 

Moreover, to the best of our knowledge there is no works that relates the musical note (understood as part of the symbolic notation) to the corresponding musical object (understood as the realisation of a musical note). In addition, none of the proposed ontologies seem to model the physical information related to the audio signal.

Therefore, this paper proposes an ontology that can represent a note both from a symbolic and a performance point of view. To do this, it is necessary to take into account both the symbolic characteristics (the signs and attributes typical of musical notation), and the physical characteristics of the note as reproduced by a musical instrument. However, this work is not intended to go into the perceptual characteristics of music. 

The problem of modelling information with respect to its realisation has already been analysed in the past. For example, the Ontology Design Pattern (ODP) \emph{Information Realization\footnote{\url{http://www.ontologydesignpatterns.org/cp/owl/informationrealization.owl}}} --extracted from DOLCE+DnS Ultralite Ontology\footnote{\url{http://ontologydesignpatterns.org/wiki/Ontology:DOLCE+DnS_Ultralite}} \cite{mascardi2007}-- proposes to help ontology designers to model this type of scenario. This pattern allows designers to model information objects and their realizations. This allows to reason about physical objects and the information they realize, by keeping them distinguished.

The same kind of relationship can be found in the FRBR\footnote{\url{https://www.ifla.org/publications/functional-requirements-for-bibliographic-records}} \cite{weiss2007} conceptual entity-relationship model, which groups entities into four different levels (work, expression, manifestation and item).  The relationship between \emph{expression}, defined as the ``specific intellectual or artistic form that a work takes each time it is realized" and \emph{manifestation}, defined as ``the physical embodiment of an expression of a work'', can be traced back to the modelling problem discussed in this paper. 

However, these examples do not solve the modelling problems that have been advanced in the previous sections. In particular, the relationship between the approximate features represented by the symbolic notation and their realisation remains a challenge in the field of ontology design. 
To achieve this, this research proposes a new ontology to represent symbolic musical notation and its realisation, considering the relationships between the abstraction of musical features expressed in the symbolic notation and the physical characteristics of the audio signal. Furthermore, this paper proposes a pattern-based approach, by proposing an ontology composed of modular and reusable elements for the representation of both audio and symbolic musical content. 

\section{The Music Note Ontology}

\begin{figure}[!ht]
    \centering
    \includegraphics[width=\textwidth]{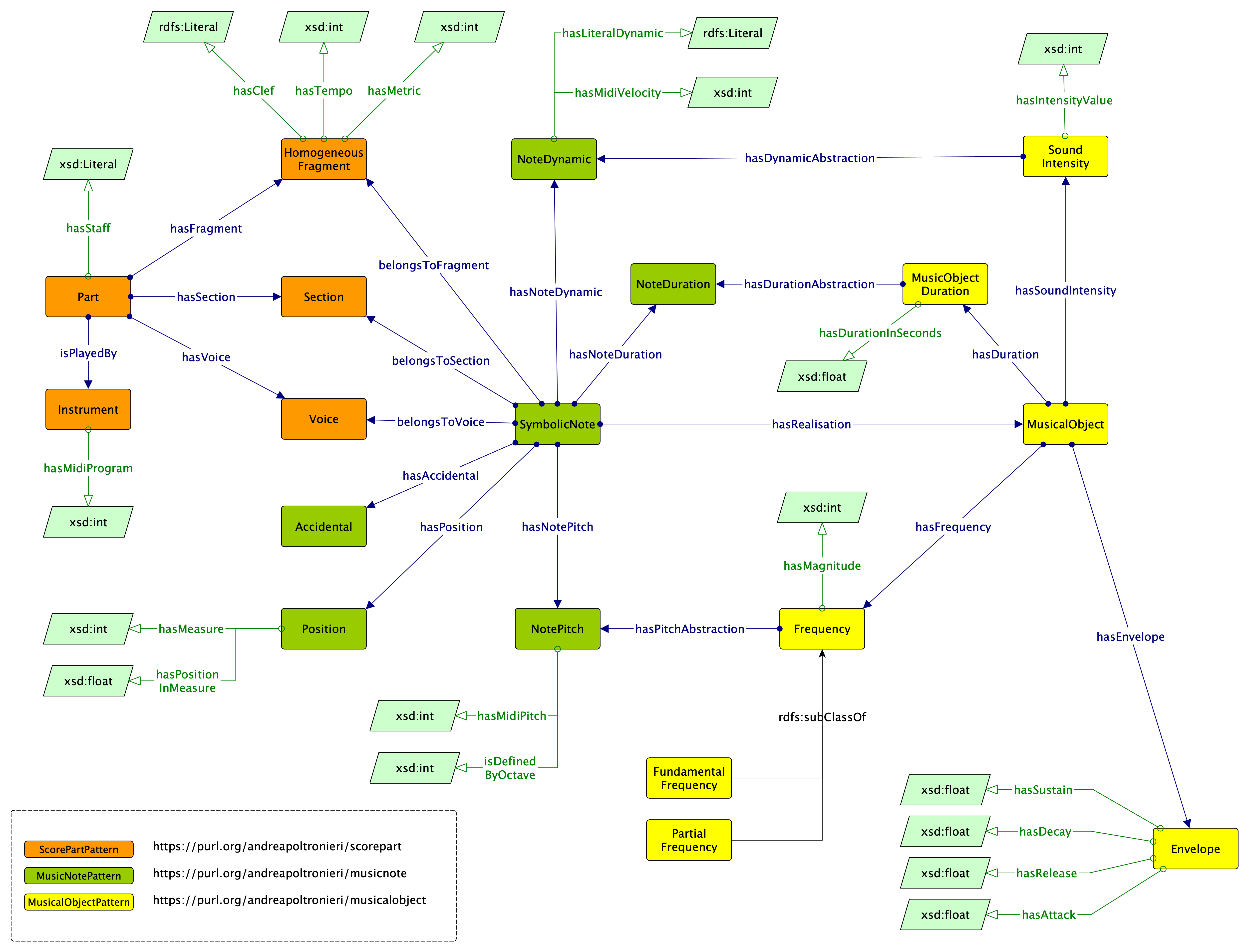}
    \caption{The Music Note Ontology in Graffoo notation. Colours define the different ODPs that constitute it.}
    \label{fig:Music note pattern}
\end{figure}

The Music Note Ontology\footnote{The OWL file of the ontology can be found at: \url{https://purl.org/andreapoltronieri/notepattern}} (see Figure \ref{fig:Music note pattern}) models a musical note both as a constituent element of the score, and as a musical object, i.e. as a realisation of the event represented by a score or symbolic notation. To do this, the proposed ontology represents both the elements and the typical hierarchies of a score, describing all its attributes. In addition, the ontology models the realisation of a note, describing its physical characteristics in a given performance. 

The proposed ontology is composed of three distinct Ontology Design Patterns, which are presented in this paper and have been submitted to the \emph{ODP portal} at the following links::
\begin{itemize}
    \item \emph{The Score Part Pattern}: \url{http://ontologydesignpatterns.org/wiki/Submissions:Scorepart} 
    \item \emph{The Musical Object Pattern}: \url{http://ontologydesignpatterns.org/wiki/Submissions:Musicalobject}
    \item \emph{The Music Note Pattern}: \url{http://ontologydesignpatterns.org/wiki/Submissions:Notepattern}.
\end{itemize}

\begin{table}
\caption{List of competency questions for the Music Note Ontology.}\label{tab1}
\begin{tabular}{ p{3.5em}  l }
\hline
ID &  Competency Question\\ 
\hline
CQ1 & what is the name of a note?\\
CQ2 & what part of the score does a note belong to?\\
CQ3 & what are the dynamic indications referring to a note in the score?\\
CQ4 & what is the fundamental frequency of a note?\\
CQ5 & what are the different frequencies that make up the spectrum of a note?\\
CQ6 & what is the duration in seconds of a note, in a given performance?\\
CQ7 & how is the envelope of a note shaped?\\
\hline
\end{tabular}
\end{table}

The note as represented in the score is described by the \emph{Music Note Pattern} (in green in the figure), which in turn imports the \emph{Score Part pattern} (in orange in the figure) and the \emph{Musical Object pattern} (in yellow in the figure). The former import describes the relationships between the different components of a score, while the latter models the musical object from the point of view of its physical features. 

A selection of competency questions that the ontology must be able to answer are listed in Table \ref{tab1}. However, the complete list of compency questions in available on the repository of the project\footnote{The project repository is available at: \url{https://github.com/andreamust/music_note_pattern/}}.

Some of the classes and properties of the ontology have also been aligned with currently available ontologies and the alignments are available in a separate file\footnote{Alignments to the Music Note Ontology are available at the URI: \url{purl.org/andreapoltronieri/notepattern-aligns}}. Alignments with other ontologies were mainly made on the Score Part Pattern and the Music Note Pattern. Although some of the available ontologies (see Section 3) define some of the classes and properties present in these patterns, for this we needed to re-model the score and symbolic elements. In fact, the objective of this work is, among others, to abstract from a specific representation system, combining the attributes of the musical score with other elements of the MIDI representation (as the most widely used symbolic representation) and, to the beat of our knowledge, there are no ontologies available with this characteristics. 

\subsection{The Score Part Pattern}

The hierarchy (i.e. the mereological structure) of a music score is described by the Score Part Pattern\footnote{Score Part Pattern URI: \url{https://purl.org/andreapoltronieri/scorepart}}.
The \texttt{Part} class describes the instrumental part of a score, which refers to a specific staff of the musical notation that is associated with an instrument. The instrument playing the part is modelled by the class \texttt{Instrument}, which is associated by means of a datatype property with the MIDI program assigned to that instrument (\texttt{hasMidiProgram}). Each part of the score is also divided into sections (class \texttt{Section}), similar music fragments (class \texttt{HomogeneousFragment}) and voices (class \texttt{Voice}). A voice is defined as one of several parts that can be performed within the same staff (e.g. \emph{alto} and \emph{soprano}). A section is instead a part of the music score defined by the repetition signs.  A fragment, on the other hand, is a grouping of notes that share the same metric, tempo and clef, described by the object properties \texttt{hasMetric}, \texttt{hasTempo} and \texttt{hasClef}.

\subsection{The Musical Object Pattern}

The Musical Object Pattern\footnote{Musical Object Pattern URI: \url{https://purl.org/andreapoltronieri/musicalobject}} models the physical features of a musical note object, i.e. the execution of a note. Specifically, this pattern models the physical characteristics that can be extracted from the sound wave produced by an instrument playing a musical note. The \texttt{MusicalObject} class is connected to four classes that describe these physical characteristics, namely duration, sound intensity, frequency and envelope. 
The \texttt{MusicObjectDuration} class expresses the duration in seconds of the musical object, by means of the object property \texttt{hasDurationInSeconds}. In the same way, the musical intensity is modelled via the \texttt{SoundIntensity} class.
Frequency is modelled by means of the class \texttt{Frequency} and its sub-classes \texttt{FundamentalFrequency} and \texttt{PartialFrequency}. For each expressed frequency, the magnitude of the frequency is also indicated using the \texttt{hasFrequencyMagnitude} datatype property. Finally, the \texttt{Envelope} class is connected to four datatype properties that describe the envelope of the waveform according to the ADSR model, namely \texttt{hasAttack}, \texttt{hasSustain}, \texttt{hasDecay} and \texttt{hasRelease}. 

\subsection{The Music Note Pattern}

The Music Note Pattern\footnote{Music Note Pattern URI: \url{https://purl.org/andreapoltronieri/notepattern}} models the symbolic note (as represented in a score or in most of symbolic representation systems) and its attributes. Moreover, this ODP aims to abstract over different music representation systems. This means that it is possible to represent with this pattern music that was originally encoded in other formats. To do so, the pattern proposed describes each symbolic notes both by means of the standard music score notation and by using the MIDI reference for each of the notes's attributes. This would allow the conversion between  

\noindent{\texttt{SymbolicNote} is the central class representing the note as represented in the score. This class has seven subclasses, one for each of the seven notes of the tempered system (primitive notes), plus 35 subclasses which are defined as the combination between primitives and accidentals.} The class \texttt{Accidental} (which has in turn five subclasses: \texttt{Flat}, \texttt{Sharp}, \texttt{Natural}, \texttt{DoubleFlat}, and \texttt{DoubleSharp}) allows the representation of altered notes. This allows to represent every note of the Western notation system, taking into account the function with respect to the tonality of the piece (e.g. differentiating between the notes A\# and Bb). Notice that such differentiation has an important cognitive function for the music reader (distinguishing enharmonics according to the tonality), function that is mostly irrelevant for a computer reading a midi, but very relevant for an AI learning to generate music from notation). Furthermore, the SymbolicNote class is linked to several classes and data properties, describing the note's attributes. 

\noindent{The class \texttt{Position} defines the position of the note both in relation to the score (datatype property \texttt{hasToMeasure}) and within the measure (datatype property \texttt{hasPositionInMeasure}). The class \texttt{NoteDuration} describes the symbolic duration of the note, as expressed in the music score. The \texttt{NoteDynamic} class, instead, describes the dynamics of the note with reference to both musical notation (datatype property \texttt{hasLiteralDynamic}, e.g \emph{crescendo}, \emph{vibrato}, etc.) and the MIDI standard (data property \texttt{hasMidiVelocity}). Similarly, the \texttt{NotePitch} class defines both the octave in which the note is located (datatype property \texttt{isDefinedByOctave}) and the MIDI pitch (datatype property \texttt{hasMidiPitch}).}
In addition, the SymbolicNote class is linked to the other two patterns that make up the Music Note Ontology. The SymbolicNote belongs in fact to a Section rather than to a Voice, while it has realisation in a MusicalObject. Finally, the relationships between the attributes of the SymbolicNote and those of the MusicalObject are expressed. 

The NotePitch of a symbolic note is represented as the abstraction of the musical object's frequency (expressed by the object property \\ \texttt{hasFrequencyAbstraction}), while NoteDynamic and NoteDuration are described as the abstraction of SoundIntensity and MusicObjectDuration, respectively (object properties \texttt{hasDynamicAbstraction} and \texttt{hasDurationAbstraction}).

In order to test the ontology, a Knowledge Base (KB)\footnote{\url{https://purl.org/andreapoltronieri/notepatterndata}} containing a single note and a single musical object was created. SPARQL queries were also created for each of the competency questions defined in Table \ref{tab1}. The KB can be tested against the SPARQL queries, in order to verify the expressiveness of the ontology with respect to the competency questions. The complete list of SPARQL queries is available on the repository of the project. 

The ontology can also be used to enrich the description of music content that is already annotated using other ontological models. For example, the classes and properties describing the structure of musical notation can be aligned to the widely used Music Ontology \cite{raimond2007}. Similarly, the same classes can be aligned with the aforementioned Music OWL and Music Score Ontology.

\section{Conclusions and Future Perspectives}

In this paper we have proposed the Music Note Ontology, an ontology for modelling musical content both from a symbolic point of view and in terms of its realisation. To do this, three different Ontology Design Patterns have been designed to model the internal relations of the score, the note from the symbolic point of view, and the note as musical object. The relationships between the different musical notations have also been considered, in particular between the symbolic abstraction and the features that can be extracted from the audio signal of a performance. 

In our future work we aim at modelling the perception of the different components of the audio signal, describing the different perceptual levels at which it is possible to experience music. We also aim to formalise the musical perception of different features from a subjective point of view, e.g. in individuals with different musical skills or from different musical cultures. Furthermore, since the model proposed at the moment can only express notes related to the traditional notation system (and thus related to the temperate system of the Western musical tradition) we plan to expand the ontology with notes from other temperaments and musical traditions, as well as unpitched notes and microtonal variations.

%
%
%
\bibliographystyle{splncs04}
\bibliography{references.bib}
%
\end{document}